\PassOptionsToPackage{table,dvipsnames}{xcolor}
\documentclass[10pt,twocolumn,letterpaper]{article}

\usepackage[pagenumbers]{cvpr}             
\usepackage{booktabs}    
\usepackage{siunitx}     
\usepackage[table]{xcolor}
\usepackage{makecell}    

\definecolor{cvprgray}{gray}{0.9}

\renewcommand{\arraystretch}{0.9}

\usepackage{array}
\newcolumntype{L}[1]{>{\raggedright\let\newline\\\arraybackslash\hspace{0pt}}m{#1}}
\newcolumntype{C}[1]{>{\centering\let\newline\\\arraybackslash\hspace{0pt}}m{#1}}
\newcolumntype{R}[1]{>{\raggedleft\let\newline\\\arraybackslash\hspace{0pt}}m{#1}}

\usepackage{multirow}
\usepackage[usestackEOL]{stackengine}
\usepackage[dvipsnames]{xcolor}

\usepackage{wrapfig} 

\definecolor{cvprblue}{rgb}{0.21,0.49,0.74}
\usepackage[pagebackref,breaklinks,colorlinks,allcolors=cvprblue]{hyperref}
\usepackage{array}
\usepackage{makecell}
\usepackage{tcolorbox}

\usepackage{booktabs}
\usepackage{multirow}
\usepackage{graphicx, amsmath, amssymb, caption, subcaption, overpic, textpos}
\usepackage{xcolor}
\usepackage{colortbl} 
\usepackage{algorithm}
\usepackage{algorithmic}

\definecolor{baselinecolor}{gray}{0.9}
\definecolor{darkgray}{gray}{0.8}

\newlength\savewidth\newcommand\shline{\noalign{\global\savewidth\arrayrulewidth
  \global\arrayrulewidth 1pt}\hline\noalign{\global\arrayrulewidth\savewidth}}
\newcommand{\tablestyle}[2]{\setlength{\tabcolsep}{#1}\renewcommand{\arraystretch}{#2}\centering\footnotesize}

\newcommand{\bI}{\mathbf{I}}

\newcommand{\bm}{\mathbf{m}}

\newcommand{\bv}{\mathbf{v}}

\newcommand{\bW}{\mathbf{W}}
\newcommand{\bx}{\mathbf{x}}

\newcommand{\bz}{\mathbf{z}}

\newcommand{\bepsilon}{{\boldsymbol{\epsilon}}}

\usepackage{orcidlink}

\usepackage{marvosym, ifsym}
\def\cvprfinalcopy  
\def\confName{CVPR}
\def\confYear{2025}

\title{TRACE: Temporally Reliable Anatomically-Conditioned 3D CT Generation with Enhanced Efficiency}

\author{%
  Minye Shao$^{1}$\orcidlink{0009-0008-8658-2174}, 
  Xingyu Miao$^{1}$\orcidlink{0000-0003-1203-8279}, 
  Haoran Duan$^{2}$\orcidlink{0000-0001-9956-7020}, 
  Zeyu Wang$^{3}$\orcidlink{0000-0002-1223-1661}, 
  Jingkun Chen$^{4}$, \\[1ex]
  Yawen Huang$^{5}$, 
  Xian Wu$^{5}$, 
  Jingjing Deng$^{6}$\orcidlink{0000-0001-9274-651X}, 
  Yang Long$^{1*}$\orcidlink{0000-0002-2445-6112}, 
  Yefeng Zheng$^{7}$\orcidlink{0000-0003-2195-2847}\\[1ex]
  $^{1}$Department of Computer Science, Durham University, Durham, UK;\\
  $^{2}$Department of Automation, Tsinghua University, Beijing, China;\\
  $^{3}$College of Computer Science and Engineering, Dalian Minzu University, Dalian, China;\\
  $^{4}$Department of Engineering Science, University of Oxford, Oxford, UK;\\
  $^{5}$Jarvis Research Center, Tencent YouTu Lab, Shenzhen, China;\\
  $^{6}$School of Engineering Mathematics and Technology, University of Bristol, Bristol, UK;\\
  $^{7}$ Medical Artificial Intelligence Laboratory, School of Engineering, Westlake University,  Hangzhou, China\\[1ex]
  {\tt\small yang.long@ieee.org}
  \thanks{Corresponding author}
}

\begin{document}
\maketitle

\begin{abstract}
3D medical image generation is essential for data augmentation and patient privacy, calling for reliable and efficient models suited for clinical practice. However, current methods suffer from limited anatomical fidelity, restricted axial length, and substantial computational cost, placing them beyond reach for regions with limited resources and infrastructure. We introduce \textbf{TRACE}, a framework that generates 3D medical images with spatiotemporal alignment using a 2D multimodal-conditioned diffusion approach. TRACE models sequential 2D slices as video frame pairs, combining segmentation priors and radiology reports for anatomical alignment, incorporating optical flow to sustain temporal coherence. During inference, an overlapping-frame strategy links frame pairs into a \textbf{flexible length} sequence, reconstructed into a spatiotemporally and anatomically aligned 3D volume. Experimental results demonstrate that TRACE effectively balances \textbf{computational efficiency} with preserving \textbf{anatomical fidelity} and spatiotemporal consistency. Code is available at: {\hypersetup{urlcolor=black}\href{https://github.com/VinyehShaw/TRACE.git}{https://github.com/VinyehShaw/TRACE}}.
\end{abstract}

\vspace{-5mm}
\section{Introduction}
\label{sec:intro}

\begin{figure}[t]
\centering
\vspace{-1mm}
\includegraphics[height=0.475\columnwidth, width=0.65\columnwidth]{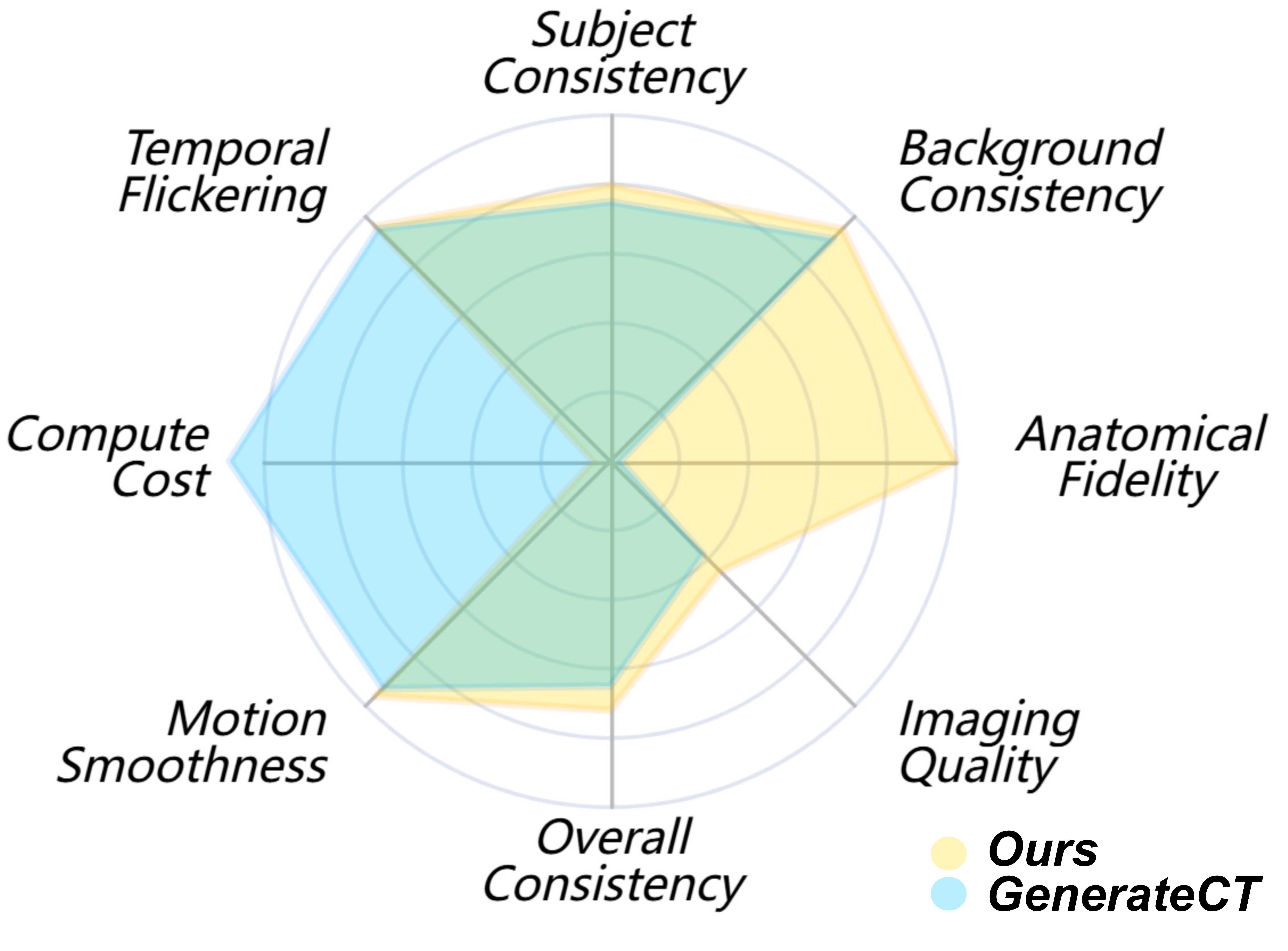}
\vspace{-4mm}
\caption{\small TRACE delivers \textbf{Flexible-Length} CT volumes, greatly cuts \textbf{Compute Cost}, boosts \textbf{Anatomical Fidelity}.}
\label{fig:trace_radar}
\vspace{-3mm}
\end{figure}
3D medical image generation is essential for precise diagnosis, personalized treatment, and surgical planning. However, privacy constraints and limited data sharing hinder the development of efficient generative models for medical imaging \cite{antonelli2022medical}. The clinical demand for high-quality, privacy-preserving diagnostic support has led to the exploration of generative methods that can effectively learn from small datasets \cite{kazerouni2023diffusion}. This approach not only addresses data scarcity but also optimizes computational resources to enhance diagnostic precision and personalized care \cite{liu20243d}.
3D diffusion models have demonstrated potential in medical image generation by preserving data privacy. To simplify model design and enhance computational efficiency, most implementations assume isotropic generation dimensions\cite{zhu2023make,han2023medgen3d}. However, 3D CT volumes often exhibit anisotropic shapes along the axial direction, with scan length and slice count adjusted according to the region of interest and pathology extent \cite{yang2020alignshift}. Consequently, the isotropic assumption in traditional 3D diffusion models may limit the precise characterization of spatial distribution features inherent to Chest CT volumes \cite{pan2023diffuseir}. Additionally, clinical applications demand high anatomical fidelity, requiring generated images to accurately capture organ shapes and details while allowing controlled variability to reflect individual differences, however, previous methods often struggle to enforce these anatomical constraints, leading to generated slices that may not faithfully represent complex anatomical structures, thus limiting clinical applicability \cite{konz2024anatomically,xu2024medsyn,chen2024general}. Finally, spatiotemporal consistency across slices is essential to maintain anatomical alignment throughout the volume, whereas achieving this continuity imposes substantial computational costs, significantly increasing training time and model complexity, which makes generating 3D CT images with both anatomical fidelity and spatiotemporal coherence a persistent challenge \cite{xu2024medsyn}.

In this paper, we propose TRACE, a novel framework for 3D CT image generation leveraging 2D diffusion models within a video generation paradigm to address the challenges of anatomical fidelity, computational cost, and spatiotemporal consistency in medical imaging. Our approach transforms 3D CT volume generation into a sequential process of 2D slice synthesis, treating each slice as a frame in a video. Additionally, we implement an overlapping frame guidance strategy during inference to ensure temporal coherence and spatial alignment across slices while avoiding the computational burden of traditional 3D models. Moreover, we introduce multimodal conditional guidance, incorporating detailed anatomical priors from segmentation masks and semantic guidance from radiology reports, to ensure anatomical accuracy in each axial slice and maintain consistent anatomical structures throughout the 3D volume, generating 3D CT images that capture realistic anatomical features with controlled variability, thereby ensuring anatomical fidelity, spatiotemporal consistency, and clinical relevance. To our knowledge, this is the first work to generate variable-length medical imaging videos using a conditionally guided 2D generation model. Our contributions are summarized as follows:
\begin{itemize}
\item
Enabling the generation of 3D CT volumes with flexible axial resolution, we propose a novel framework that models 3D medical volumes as sequences of continuous frames, leveraging multi-conditioned 2D diffusion to achieve substantial computational efficiency.
\item
Achieving anatomical fidelity and spatiotemporal consistency, our framework integrates frame skipping and positional encoding for temporal coherence, spatial alignment through optical flow, and anatomical priors derived from segmentation masks and radiology reports. This ensures precise alignment of each frame and its transitions with anatomical structures in the generated volumes.
\item 
We enhance previous evaluation methods on this task by introducing an anatomical fidelity assessment for generated volumes. Extensive experiments demonstrate our method’s robust performance on limited training data, achieving substantial improvements in anatomical fidelity while reducing training and inference costs by 87.5\% and 92.5\%, respectively.
\end{itemize} 

TRACE supports essential applications in medical imaging, including data augmentation, where anatomically fidelity synthetic images enhance training sets and improve model generalization in data-limited contexts. In privacy-preserving data sharing, TRACE enables research without compromising patient identity. For personalized medicine, clinicians can define anatomy via text and masks, aiding diagnosis, surgical planning, and training. These applications underscore TRACE's utility in expanding data resources, safeguarding privacy, and supporting clinical workflows.
\vspace{-2mm}
\section{Related Work}
\vspace{-0.5mm}
\subsection{Medical Generative Modeling}
\vspace{-0.5mm}
With the growing maturity of deep learning \cite{zhai2025dsleepnet, li2024sid,zhang2024depth,10323083,wan2024sentinel,miao2025rethinking,miao2025laser,duan2023dynamic, duan2025parameter, li2025unified, li2025bp, chang2023design,chang25largescale,chen2024hint,ZHANG2025113393,Zhang_2025_BMVC}, generative modeling has witnessed rapid advances. Recent advancements in medical generative modeling have leveraged diffusion-based methods \cite{gungor2023adaptive, jiang2023cola, lyu2022conversion, ozbey2023unsupervised, Siamese-Diffusion}, demonstrating impressive capabilities in producing refined medical images. Conditional diffusion models, specifically text-guided diffusion models \cite{chambon2022roentgen,lee2023unified}, generate medical images based on textual descriptions, aligning the outputs with specific clinical narratives and supporting applications such as data augmentation and diagnostic training.
Text-conditioned video generation, including diffusion-based and transformer-based autoregressive methods, has seen notable progress \cite{rombach2022high, ho2022imagen, ho2022video, hong2022cogvideo, villegas2022phenaki, wu2021godiva, wu2022nuwa}. Diffusion-based models typically employ 3D U-Nets to generate short, low-resolution videos with fixed frame counts, subsequently enhancing resolution and duration through cascaded techniques \cite{ho2022imagen}. In contrast, transformer-based models offer flexibility with variable frame counts and longer sequences, though often at lower resolutions \cite{yan2021videogpt}. These advances have inspired adapting video generation techniques to 3D medical imaging, treating 2D slices as sequential frames. GenerateCT \cite{hamamci2023generatect} integrates transformer and diffusion models to synthesize coherent 3D CT volumes by employing spatial and temporal attention. However, its high computational cost arises from using a causal vision transformer for 3D encoding, vision-language alignment, and cascaded diffusion for super-resolution, with each stage contributing to the model’s complexity.
\vspace{-0.5mm}
\subsection{Semantic Synthesis}
\vspace{-0.5mm}
Semantic synthesis, or generating images from segmentation masks, is a critical task in medical image generation, enabling precise control over anatomical structures in synthesized images. Traditional image-to-image translation methods, including GAN-based models \cite{cao2023deep,choi2018stargan,yang2019unsupervised} and diffusion-based models \cite{rombach2022high,wolleb2022diffusion,zhang2023adding}, have been employed for this purpose but often lack the ability to enforce strict pixel-wise anatomical constraints essential for medical applications. Recent works have attempted to fine-tune large pre-trained latent diffusion models (LDMs) for segmentation-conditioned synthesis on natural images \cite{wang2022pretraining,zhang2023adding}; however, these models adapt poorly to medical images due to the complex anatomical variations and potential loss of spatial details when converting conditioning masks into abstract latent spaces. To address these challenges, Konz et al. \cite{konz2024anatomically} proposed SegGuidedDiff, a diffusion model that conditions multi-class anatomical segmentation masks at each sampling step, ensuring images adhere closely to anatomical constraints. Similarly, LeFusion \cite{lalande2022deep} introduces a lesion-focused diffusion model that synthesizes images by concentrating on lesion areas, improving performance on cardiac MRI datasets. In histopathology image synthesis, methods such as InsMix \cite{lin2022insmix} and DiffMix \cite{oh2023diffmix}, which apply random label perturbations to generate images from modified labels, while others generate randomly distributed labels for image synthesis \cite{butte2022sharp,hou2019robust,yu2023diffusion}. Although these approaches enhance control over the synthesis process, they often focus on single-class data or involve complex two-stage diffusion processes, which may not capture the spatial and structural correlations essential for anatomically consistent 3D medical images. 
\vspace{-2mm}
\section{Methods}
\vspace{-1mm}
We propose a novel 3D CT generation approach leveraging a 2D diffusion model within a video generation framework. Our method captures diverse temporal relationships through variable interval frame pairs and ensures smooth transitions through an Overlapping Frame Guidance strategy. High-fidelity generation is achieved by integrating optical flow for temporal coherence, text prompts for semantic context, positional embeddings for temporal positioning, and anatomical priors from NVIDIA VISTA3D \cite{vista3d} for structural accuracy, resulting in anatomical fidelity and customized CT synthesis.
\begin{figure*}
    \centering
   
    \includegraphics[width=1\textwidth]{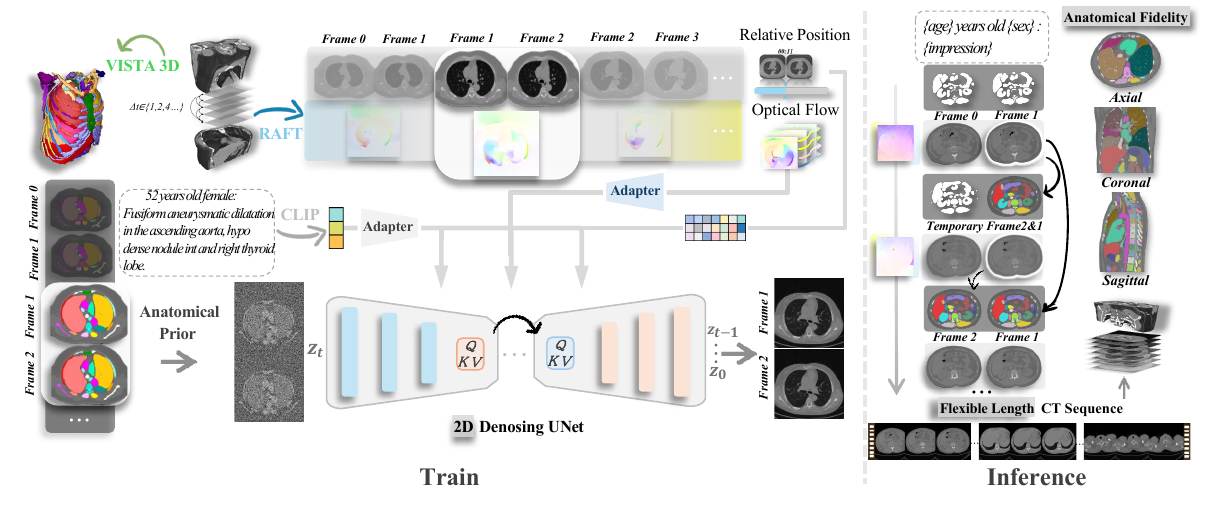}
 \vspace{-5mm}
  \caption{TRACE models 3D CT volumes as sequences of frames, utilizing an efficient 2D diffusion model conditioned on multiple modalities to generate flexible length, coherent CT sequences. During training, it denoises frame pairs with varying skip intervals, guided by four modalities: anatomical masks (VISTA3D), optical flow between frames (RAFT), report embeddings (CLIP), and relative position embeddings. The optical flow and text embeddings pass through trainable adapters before entering the diffusion model. Inference employs an overlapping-frame guidance strategy to synthesize semantically aligned frame pairs, generating anatomically consistent CT sequences, which are then reconstructed back to 3D volumes.}
   
    \label{first}
     
\end{figure*}

\subsection{Preliminaries}

\subsubsection{Diffusion Probabilistic Models (DPMs)} DPMs are generative models that approximate the data distribution by learning a reverse denoising process \cite{song2020score}. The forward process progressively adds Gaussian noise to the data, transitioning from the original data to pure noise over \( T \) steps, defined as: 
\begin{equation}
q(\bx_t \mid \bx_{t-1}) = \mathcal{N}(\sqrt{1 - \beta_t} \bx_{t-1}, \beta_t \bI),
\end{equation}
where \( \beta_t \) is a variance schedule. The reverse process learns to remove this noise step-by-step.

Denoising Diffusion Implicit Models (DDIM) enable faster sampling by modifying the forward process, allowing deterministic generation: 
\begin{equation}
\bx_{t-1} = \sqrt{\alpha_{t-1}} f_{\theta}(\bx_t, t) + \sqrt{1 - \alpha_{t-1}} \epsilon_{\theta}(\bx_t, t),
\end{equation}
where setting \( \sigma_t = 0 \) ensures consistent generation, useful for video-like applications.
\subsubsection{3D Medical Imaging as Video}
To enable flexible axial resolution, allowing the generation of 3D CT volumes with variable numbers of slices, we treat 3D medical images as a sequence of 2D slices, analogous to frames in a video. Let \( V \in \mathbb{R}^{H \times W \times D} \) represent a 3D volume, where \( H \), \( W \), and \( D \) denote height, width, and depth (number of slices), respectively. By considering the depth dimension as a temporal axis, diffusion models can effectively generate 3D data. This approach not only provides customizable axial lengths to meet diverse clinical requirements but also significantly reduces computational costs by producing only the necessary slice sequences.
\subsection{Conditional 2D Diffusion for 3D CT Synthesis}
\subsubsection{Paired Frame Temporal Modeling}
In our framework, we utilize a 2D diffusion model to generate 3D CT image sequences by operating on frame pairs. Specifically, each consecutive frame pair is concatenated along the channel dimension, forming an input tensor of shape \((B, 2C, H, W)\), where \( B \) is the batch size, \( C \) is the original number of channels (e.g., grayscale or RGB), and \( H, W \) denote the spatial dimensions. By concatenating two frames as separate channels, the model’s convolutional filters can jointly process both frames, allowing it to detect spatial-temporal patterns, such as motion or changes between frames, through standard 2D convolutions. This approach enables the model to capture temporal dynamics effectively while maintaining the computational efficiency of a 2D architecture. Unlike conventional 3D diffusion models that explicitly handle the temporal dimension, our method leverages the channel dimension to encode temporal information by treating each frame pair as correlated features. This formulation encourages the model to establish relationships between the two frames, effectively modeling temporal continuity without requiring an explicit temporal dimension.\vspace{0.1em}
\setlength{\columnsep}{1mm}%
\begin{wrapfigure}[1]{r}[0cm]{0.4\columnwidth}%
\vspace{-5.2mm}\includegraphics[width=0.4\columnwidth]{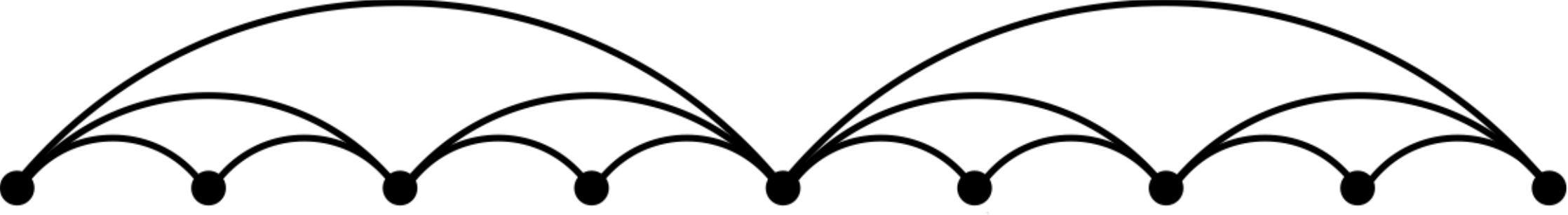}%
\end{wrapfigure}%
\hspace{1.5em}To further enhance the model's ability to capture both short and long-term spatial dependencies, we introduce a frame-skipping strategy designed to adapt to the Z-axis spacing (Z-spacing) characteristics typical of CT imaging. In real clinical settings, CT images often vary in Z-spacing depending on anatomical requirements \cite{moser2017radiation}, and in our dataset, most samples are set to spacings of 0.7, 1.5, or 3 mm. To align with this characteristic, our frame-skipping intervals are set to 1, 2, and 4, respectively, simulating the variation in Z-spacing. Formally, this strategy is defined as:
\begin{equation}
\scalebox{0.91}{$
P = \{(i,j)\mid j = i + k,\;i \bmod k = 0,\;k \in \{1,2,4\}\}.$}
\end{equation}
\noindent By training on these diverse frame pairs, the model gains a richer understanding of temporal relationships, reinforcing spatial consistency across different temporal spans and resulting in more coherent and robust outputs during inference.


For each frame pair \((i, j) \in P\), we compute a dense optical flow field \(f_{i \to j}\) (mapping a pixel in frame \(i\) to its corresponding location in frame \(j\)) using RAFT \cite{teed2020raft}. This optical flow is then incorporated into the model through a trainable adapter, providing additional guidance for temporal alignment between frames.

The learned temporal coherence can be further explained through the shared representation in the latent space. During training, the concatenated frames are treated as a joint entity, and the denoising process iteratively updates them to minimize inconsistencies. Let \(\bz_t^{(i)}\) and \(\bz_t^{(j)}\) represent the latent representations of frames \(i\) and \(j\) at time step \(t\). The temporal consistency learned during training can be interpreted as the minimization of an objective function that enforces similarity between the two frames in the latent space:

\begin{equation}
T_{\text{coherence}} = \mathbb{E}_{(i, j) \in P} \left[ ||\bx_t^{(i)} - \bx_t^{(j)}||^2 \right].
\end{equation}

\noindent This objective encourages the latent representations of the two frames to be similar, thus enforcing temporal coherence during both training and inference.

\subsubsection{Anatomical Guidance}
We condition the generation of an image \(\bx_0 \in \mathbb{R}^{c \times h \times w}\) on a multi-class anatomical mask \(\bm \in \{0, \dots, n - 1\}^{h \times w}\), where 
 \(c\)  is the number of channels, \(h\) and  \(w\) are the height and width of the image, \(n\) represents the number of anatomical classes, including background. These anatomical masks, obtained from VISTA3D with 127 classes, cover comprehensive human structures, including organs, bones, and lesions. We aim to sample from the distribution \(p(\bx_0 \mid \bm)\), ensuring the generated images adhere to the provided anatomical context.

The forward noising process \(q(\bx_t \mid \bx_{t-1})\) remains unchanged. However, the reverse process \(p_{\theta}(\bx_{t-1} \mid \bx_t, \bm)\) and the noise prediction network \(\epsilon_{\theta}\) are conditioned on the anatomical mask. This conditioning modifies the denoising process to incorporate anatomical priors effectively. The training objective is formulated as:
\begin{equation}
\scalebox{0.88}{$
L_m = \mathbb{E}_{(\bx_0, \bm), t, \bepsilon} \left[ \left\| \bepsilon - \bepsilon_{\theta}\left( \sqrt{\alpha_t} \bx_0 + \sqrt{1 - \alpha_t} \bepsilon, \, t, \bm \right) \right\|^2 \right],$}
\end{equation}
\noindent where \(x_0\) is paired with its corresponding anatomical mask \(m\) for each training instance. This objective ensures that the anatomical structures provided by the mask are respected during the generation.

For paired frame generation, anatomical masks for both frames are computed using VISTA3D and concatenated channel-wise with the network input at every denoising step, allowing the model to access anatomical information at the pixel level. Let \((\bm_i, \bm_j)\) denote the anatomical masks for frame pair \((i, j)\). The conditioning is integrated as follows:
\begin{equation}
\scalebox{1}{$
\vspace{-3mm}
\bepsilon_{\theta}(\bx_t^{(i, j)}, t \mid \bm_i, \bm_j) : \mathbb{R}^{(2c+2) \times h \times w} \to \mathbb{R}^{2c \times h \times w}.$}\vspace{-3mm}
\end{equation}
\noindent Additionally, we incorporate semantic information from textual guidance to enhance anatomical fidelity further. The text prompt includes patient-specific information such as age, gender, and diagnostic impressions, providing valuable semantic cues for the model. The text is processed using a pre-trained CLIP model to extract a text embedding \(\bv_t\), which encodes rich semantic information.

The text embedding \(v_t\) is passed through a trainable adapter designed to transform the semantic features for effective conditioning. Let \(\bW_1, \bW_2\) represent the learnable parameters of the adapter, consisting of a series of linear transformations with nonlinear activations to capture complex semantic interactions: \( v'_t:\phi_2(\bW_2 \cdot \phi_1(\bW_1 v_t))\), where \(\phi_1\) and \(\phi_2\) are nonlinear activation functions applied to introduce nonlinearity and improve expressiveness. The transformed vector \(v'_t\) is subsequently used to condition the model by integrating it into the encoder hidden states \(h_t\): \( h_t = h_t + v'_t \).

 This dual approach, combining anatomical mask guidance and textual semantic information, ensures the generated images respect both pixel-level anatomical structures and high-level semantic context, enhancing spatial accuracy and temporal coherence, crucial for medical imaging applications.

\subsubsection{Temporal Position Encoding}
To encode the temporal position of each frame pair within the entire video sequence, we employ sinusoidal positional embeddings. Given the start frame \(f_0\) and the end frame \(f_N\), the relative position \(r_i\) for frame \(f_i\) is computed as:
\begin{equation}
\scalebox{0.87}{$
E(r_i)_k = \sin\left( \frac{r_i}{10000\lambda^{2k / d}} \right),\quad E(r_i)_{k+1} = \cos\left( \frac{r_i}{10000\lambda^{2k / d}} \right),$}
\end{equation}
\noindent where \(r_i = \frac{f_i - f_0}{f_N - f_0}\) represents the normalized relative position, \(d\) is the embedding dimension, \(k\) indexes the embedding dimension, and \(\lambda\) is a scaling factor. This formulation provides a compact and continuous representation of the frame's temporal position.

These embeddings are concatenated for both frames in a pair and integrated as conditional guidance within the diffusion model: \(\bepsilon_{\theta}(\bx_t^{(i, j)}, t, E(r_i, r_j))\).

This design allows the model to adjust the denoising dynamics based on the temporal position of the frame pair, enhancing temporal coherence throughout the generated sequence. By embedding temporal position implicitly into the conditioning, the model better understands the relative context, leading to smoother transitions and consistent temporal alignment in the generated CT images.

\subsection{Inference via Overlapping Frame Guidance}
Traditional diffusion models, often formulated as Markovian processes, result in a ``memoryless'' progression, where each generation step depends solely on the previous state. However, as mentioned, thoracic CT imaging typically has a relatively small Z-axis spacing to allow detailed observation of fine anatomical structures, leading to a characteristic of approximately consistent and subtle changes between frames. Similarly, generated CT images require continuity and coherence across time and frames. To address this need and align with these characteristics, we propose the overlapping frame guidance strategy, a non-Markovian approach where each generated frame directly contributes to forming subsequent frames. Through this layered guidance strategy, our method extends anatomical coherence across the entire sequence, allowing each frame to shape the next in a controlled and continuous manner, thereby fulfilling the specific requirements of CT imaging.

As shown in \cref{algo1}, the inference process begins by generating the initial frame pair $(\bx(0), \bx(1))$, relying on the anatomical priors $M(0)$ and $M(1)$. These segmentation masks serve as essential guides, structuring the initial input as $\tilde{\bx}_t = \text{Concat}(\bx_t, M(0), M(1))$. This input, combining random noise with anatomical context, is passed through the DDIM process to produce the initial frames $(\bx(0), \bx(1))$ that anchor the generation sequence.

For each subsequent frame pair, the process leverages overlapping frame guidance to ensure continuity. Specifically, for frame $\bx(n+1)$, the previously generated frame $\bx(n)$ and its mask $M(n)$ are incorporated to create a guidance map $G(n)$. This map is constructed by processing $\bx(n)$ through a transformation $\hat{\bx}(n) = \mathcal{F}\left( \mathcal{H}\left(\bx(n)^\gamma \right) \right)$, where $\gamma$ amplifies high-intensity features, $\mathcal{H}$ selectively smooths the background, and $\mathcal{F}$ normalizes the values. This results in $G(n) = M(n) + [1 - M(n)] \cdot \hat{\bx}(n)$, a map that emphasizes anatomical regions while smoothing transitions elsewhere.

The generated guidance map $G(n)$, together with the segmentation mask $M(n+1)$, forms the input $\tilde{\bx}_t=\text{Concat}(\bx_t, G(n), M(n+1))$ for the next DDIM step, producing frames $(\tilde{\bx}(n), \tilde{\bx}(n+1))$. This recursive setup, detailed in the pseudocode, ensures each new frame is conditioned on the prior frame, promoting temporal coherence across the sequence \( \{\bx(n) \}_{n=0}^{N} \) while preserving anatomical fidelity.

Through overlapping frame guidance, each frame sequentially shapes the next, creating a cascade of context-driven synthesis that aligns well with anatomical structures and segmentation masks, achieving coherence and continuity throughout the sequence.
\setlength{\textfloatsep}{5pt}
\begin{algorithm}[t]
\caption{Overlapping Frame Guidance Inference}
\label{algo1}
\begin{algorithmic}[1]
\STATE \textbf{Given:} Total frames $N$, Anatomical Prior $\{M(n)\}_{n=0}^{N}$
\vspace{-4.4mm}
\STATE \textbf{Initialize:}
\STATE $\tilde{\bx}_t \gets \text{Concat}(x_t,\ M(0),\ M(1))$
\STATE $(\bx(0),\ \bx(1)) \gets \text{DDIM}(\tilde{\bx}_t)$
\FOR{$n = 1$ to $N-1$}
    \STATE $G(n) \gets M(n) + [1 - M(n)] \cdot \mathcal{F}\left( \mathcal{H}\left( \bx(n)^\gamma \right) \right)$
    \STATE $\tilde{\bx}_t \gets \text{Concat}(\bx_t,\ G(n),\ M(n+1))$
    \STATE $(\tilde{\bx}(n),\ \tilde{\bx}(n+1)) \gets \text{DDIM}(\tilde{\bx}_t)$
    \STATE $G(n+1) \gets M(n+1) + [1 - M(n+1)] \cdot \mathcal{F}\left( \mathcal{H}\left( \tilde{\bx}(n)^\gamma \right) \right)$
    \STATE $\tilde{\bx}_t \gets \text{Concat}(\bx_t,\ G(n),\ G(n+1))$
    \STATE $(\bx(n),\ \bx(n+1)) \gets \text{DDIM}(\tilde{\bx}_t)$
\ENDFOR
\STATE \textbf{Output:} Frames $\{\bx(n)\}_{n=0}^{N}$
\end{algorithmic}
\end{algorithm}

\begin{table*}[htbp]
\centering
\setlength\tabcolsep{3pt}
\small 
\vspace{-10pt}
\caption{Comparison of quantitative and ablation experiments evaluates scores between our method and GenerateCT across six VBench dimensions. Segmentation results of generated volumes are assessed using three common medical segmentation metrics, serving as reference indicators for anatomical fidelity evaluation.}
\vspace{-2mm}
\resizebox{0.9\linewidth}{!}{%
\begin{tabular}{c|c|c|c|c|c|c|c|c|c|c|c}
\toprule
\rowcolor{darkgray}
\textbf{Method}   & \textbf{\Centerstack{Subject\\Consistency}} & \textbf{\Centerstack{Background\\Consistency}} & 
\textbf{\Centerstack{Temporal\\Flickering}} & \textbf{\Centerstack{Motion\\Smoothness}} & \textbf{\Centerstack{Imaging\\Quality}} & \textbf{\Centerstack{Overall\\Consistency}} &  
\textbf{\Centerstack{JI}}$\uparrow$ & \textbf{\Centerstack{DC}}$\uparrow$  & \textbf{\Centerstack{95HD}}$\downarrow$ & \textbf{\Centerstack{FPS}}$\uparrow$ & \textbf{\Centerstack{Train/Infer\\VRAM}} \\ 
\Xhline{1pt}
GT                 & 84.58\% & 95.26\% & 99.08\% & 99.41\% & 52.89\% & 21.58\% & 100\% & 100\% & 0.00 & –    & –          \\  
\midrule
GenerateCT~\cite{hamamci2023generatect}
                   & 77.78\% & 94.73\% & 94.01\% & 95.50\% & 48.56\% & 19.58\% 
                   & 4.98\% & 9.49\% & 220.07 & 1.09 & 8×80GB/80GB \\
MedSyn~\cite{xu2024medsyn}           
                   & 74.30\% & 91.70\% & 95.60\% & 95.70\% & 50.30\% & 20.80\% 
                   & 35.40\% & 47.80\% & 59.90  & 3.91 & 4×48GB/96GB \\
\midrule
\rowcolor{baselinecolor}
PFM                & 67.30\% & 90.70\% & 94.80\% & 73.20\% & 32.70\% & 15.30\% 
                   & 20.90\% & 25.30\% & 87.40  & \textbf{6.53} & 80GB/6GB   \\
\rowcolor{baselinecolor}
DAG                & 54.50\% & 86.40\% & 85.70\% & 90.20\% & 50.30\% & 18.50\% 
                   & 52.70\% & 59.40\% & 33.20  & 5.52 & 80GB/6GB   \\
\rowcolor{baselinecolor}
PFM, DAG           & 67.70\% & 92.80\% & 95.50\% & 96.20\% & 51.20\% & 19.10\% 
                   & 54.60\% & 63.50\% & 30.90  & 4.98 & 80GB/6GB   \\
\midrule
\rowcolor{baselinecolor}
50/[1]             & 77.31\% & 92.93\% & 93.37\% & 94.82\% & 51.46\% & 21.02\% 
                   & 58.78\% & 73.43\% & 19.18  & 4.27 & 80GB/6GB   \\
\rowcolor{baselinecolor}
100/[1,2]          & 78.25\% & 94.61\% & 95.32\% & 96.04\% & 50.32\% & 19.78\% 
                   & 59.76\% & 69.38\% & 21.80  & 3.56 & 80GB/6GB   \\
\rowcolor{baselinecolor}
50/[1,2,4]        & 80.07\% & 95.08\% & 95.98\% & \textbf{97.29}\% & \textbf{53.91}\% & 19.33\% 
                   & 62.77\% & 73.48\% & 15.25  & 3.89 & 80GB/6GB   \\
\rowcolor{baselinecolor}
\textbf{100/[1,2,4] (Ours)}  
                   & \textbf{80.13}\% & \textbf{95.17}\% & \textbf{96.07}\% & 96.69\% & 52.58\% & \textbf{21.63}\% 
                   & \textbf{62.94}\% & \textbf{76.50}\% & \textbf{13.37} & 3.45 & 80GB/6GB \\ 
\Xhline{1pt}
\end{tabular}%
}

\label{tab1}
\end{table*}

\section{Experiment and Results}

\subsection{Dataset and Implementation Details}
\textbf{Dataset.} We utilized the CT-RATE \cite{hamamci2024foundation} dataset, comprising de-identified chest CT volumes and radiology reports from 21,314 patients under diverse imaging conditions \cite{lamba2014ct}. The training set consisted of 100 randomly selected CT volumes (512 $\times$ 512 pixels, 150–600 slices each), totaling approximately 22,000 scans. The test sets consist solely of volumes from unseen patients. We performed Hounsfield Unit (HU) value recalibration and standardized the spatial orientation and voxel spacing (X, Y, Z intervals) of the CT volumes to ensure consistency in image density and anatomical spatial accuracy. Radiology reports were parsed into text prompts formatted as ``\{\textit{age}\} years old \{\textit{sex}\}: \{\textit{impression}\}''. 

\noindent \textbf{Implementation Details.} We employed the AdamW optimizer \cite{loshchilov2017decoupled} with an initial learning rate of $1 \times 10^{-5}$ and a cosine annealing schedule (35,000 warmup steps, minimum learning rate $1 \times 10^{-6}$). Training spanned 300 epochs with a batch size of 28, taking approximately 14 days on a single NVIDIA A100 GPU. CT slices were downsampled to $256 \times 256$ pixels for memory efficiency, which also matches the resolution of generated images and segmentation maps. Frame pairs used skip intervals of [1, 2, 4] to enhance temporal diversity, while optical flow was computed at the original resolution. A frame embedding dimension of 64 was utilized. During inference, a batch size of 1 enabled efficient processing on GPUs with 6 GB of memory.

\begin{figure*}
    \centering
    \label{first}
    \includegraphics[width=1\textwidth]{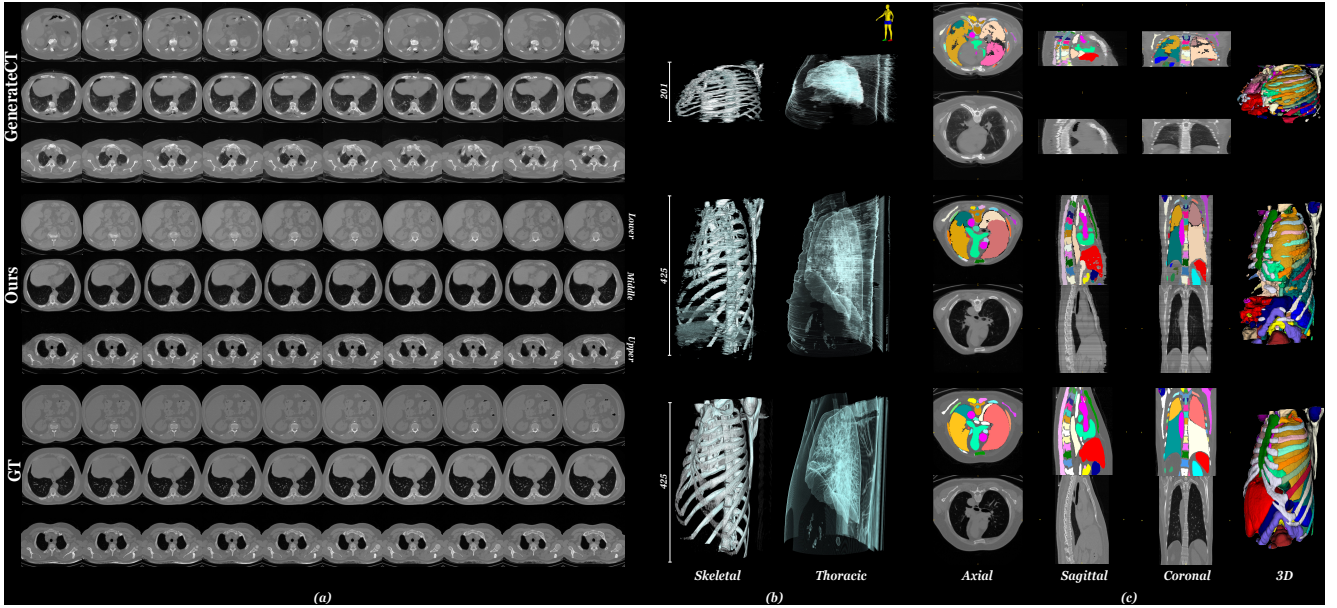}
\vspace{-5mm}
\caption{Comparison of generated results from multiple perspectives for ``52 years old male: Fusiform dilatation in the thoracic aorta. Hepatomegaly, hepatosteatosis. Hiatal hernia. Hypodense nodule in the right thyroid lobe.'' (a) Comparison of axial slices from GenerateCT, our method, and ground truth (GT), arranged left to right from diaphragm to clavicle, with each method displaying the upper, middle, and lower thorax (frames 9-18, 174-183, and 379-388). (b) 3D rendering comparison highlighting the skeleton, thoracic cavity, and key lung structures. (c) Segmentation results on generated volumes in axial, sagittal, and coronal views, with corresponding 3D renderings.}
\label{fig3}
\vspace{-3mm} 
\end{figure*}
\subsection{Quantitative Evaluation}
Building on GenerateCT, we introduce an optimized evaluation framework that improves assessments of temporal consistency, frame quality, semantic alignment, and memory efficiency, adding segmentation metrics to measure anatomical fidelity, see \cref{tab1}.

\noindent \textbf{Anatomical Fidelity.} Generated 3D volumes are standardized to HU with uniform orientation and voxel spacing, then segmented with VISTA3D \cite{vista3d}. We quantify fidelity by comparing generated and ground truth segmentations using Dice Coefficient (DC), Jaccard Index (JI), and 95th percentile Hausdorff Distance (95HD) \cite{JI}.

\noindent \textbf{VBench.} VBench \cite{huang2024vbench} evaluates key video aspects: Subject Consistency (DINO \cite{dino}), Background Consistency (CLIP \cite{clip}), Temporal Flickering, Motion Smoothness (video interpolation priors \cite{li2023amt}), Imaging Quality (MUSIQ \cite{ke2021musiq} on SPAQ \cite{fang2020perceptual}), and Overall Consistency (ViCLIP \cite{viclip}), all metrics scoring higher indicating better result. This framework precisely assesses CT sequence continuity and consistency.

\noindent \textbf{FVD\&FID.} Fréchet Inception Distance (FID) measures slice-level fidelity, while Fréchet Video Distance (FVD) extends FID to video, assessing quality and temporal coherence using I3D \cite{FVD}. Lower distances reflect better fidelity in medical image generation \cite{heusel2017gans}.

As shown in \cref{tab1}, our method outperforms GenerateCT and MedSyn across benchmark video evaluation metrics, notably in \textbf{Subject Consistency} and \textbf{Motion Smoothness}, demonstrating superior frame continuity and smooth motion dynamics. For anatomical fidelity, our approach achieves Dice and Jaccard scores that surpass GenerateCT by 1163\% and 706\%, respectively, indicating substantial improvements in anatomical accuracy. FPS reflects the inference throughput. Since our method generates 2D frame pairs, we report the average slices per second at the CT volume level for fair comparison across methods, following the official settings and using a single NVIDIA H20 GPU (96GB) for all methods. Regarding computational efficiency,
TRACE only requires a GTX 1660 Ti (6GB) for inference and one A100 (80GB) for training, while GenerateCT needs 8 A100s (80GB each) for training and one A100 for inference, highlighting our efficiency advantage.

\subsection{Qualitative Results}
To demonstrate the effectiveness of TRACE in aligning with the requirements of clinical 3D medical image generation, we designed qualitative experiments focusing on spatiotemporal and anatomical consistency when generating sequences of flexible lengths.

\noindent\textbf{Spatiotemporal Consistency.} \cref{fig3} (a) illustrates the temporal coherence of the generated sequences and the adaptability of TRACE in producing volumes of variable lengths. We extracted multiple axial slices from three key anatomical regions, extending from the diaphragm to the clavicle, including the lower thorax (lung base), middle thorax (lung hilum), and upper thorax (lung apex). As seen in the middle row of \cref{fig3} (a), TRACE demonstrates smooth transitions across frames, capturing subtle anatomical variations while maintaining spatial consistency throughout the volume. This consistency is evident in the continuity of anatomical structures such as ribs, vertebrae, and lung fields across adjacent slices. In contrast, GenerateCT exhibits noticeable discontinuities and inconsistencies between frames, with abrupt anatomical changes that may impede clinical interpretation. Furthermore, TRACE generates volumes of arbitrary lengths without sacrificing temporal coherence, an essential feature in medical imaging, where scan lengths vary according to diagnostic requirements. The consistent quality across sequences of different lengths highlights the scalability and robustness of our approach.

\noindent\textbf{Anatomical  Fidelity.}
We evaluate anatomical fidelity from both orthogonal planes and a 3D perspective. Segmentation results are derived from the VISTA3D segmentation model applied to the generated volumes, allowing for anatomical consistency assessment through key organ and skeletal structure recognition. The generated volumes were standardized to HU with aligned spatial orientation and voxel spacing, using DICOM metadata parameters from the ground truth, ensuring a consistent basis for comparison. For 2D evaluation (\cref{fig3} (c)), we extract and compare the segmentation and original images from Axial, Sagittal, and Coronal views generated by TRACE and GenerateCT. TRACE accurately represents anatomical structures and pathologies, maintaining appropriate tissue contrast and correct anatomical positioning, ensuring slice-by-slice consistency. In contrast, GenerateCT may present distortions or misaligned anatomical features, with organs or skeletal structures incorrectly positioned or missing, which limits its diagnostic relevance. For 3D evaluation, as shown in \cref{fig3} (b), TRACE produces detailed and spatially coherent renderings of thoracic anatomy, including the lung parenchyma, airways, rib cage, and vertebral column, while retaining the original axial length. In contrast, GenerateCT is limited to 201 axial slices compared to 425 in the original volume, resulting in significant information loss. Figure \cref{fig3} (c-3D) further contrasts the 3D segmentations of the thoracic region, showing that TRACE preserves known anatomical structures effectively, whereas GenerateCT's reduced axial resolution and positional inconsistencies lead to less reliable segmentation, most notably with complete omission of structures like the sternum, small intestine, and colon.

\subsection{Additional Comparison Study}
\noindent \textbf{Existing Methods.} We selected four state-of-the-art methods: Imagen \cite{saharia2022photorealistic}, Stable Diffusion (SD) \cite{rombach2022high}, Phenaki \cite{villegas2022phenaki}, and GenerateCT \cite{hamamci2023generatect}. Imagen is a 2D image generation method conditioned on text prompts, generating high-resolution medical images slice by slice based on slice indices and textual cues. Stable Diffusion is a pre-trained 2D text-to-image model fine-tuned for medical image generation to assess its performance in 3D generation tasks. Phenaki, an advanced text-to-video generation model, was adapted for 3D chest CT image generation to evaluate its effectiveness in this domain. GenerateCT provided reference results for fine-tuning and experimentation with the above methods as the first framework to generate high-resolution 3D chest CT images based on natural language prompts and our previous work.

\begin{figure}

    \centering

    \includegraphics[width=1\columnwidth]{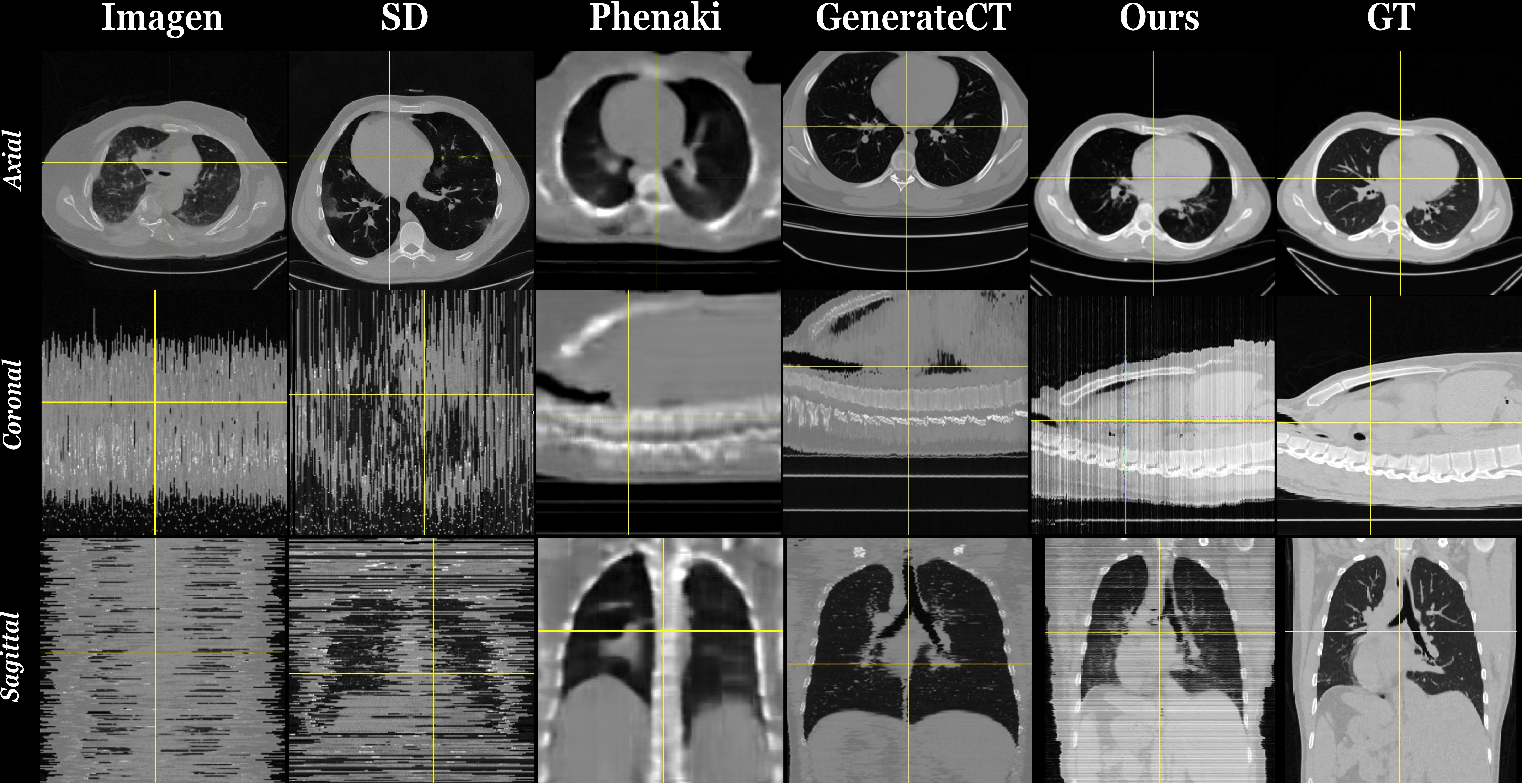}
\vspace{-7mm}
    \caption{Axial, sagittal, and coronal slices of 3D CT volumes generated by various methods, case: ``26-year-old male: Findings compatible with COVID-19 pneumonia''.}
    \label{fig4}

\end{figure}

\noindent \textbf{Results and Effectiveness.} Compared to Imagen and Stable Diffusion, as shown in \cref{fig4}, the coronal and sagittal planes reflect superior spatiotemporal coherence in our method. Leveraging optical flow-guided frame skipping and overlapping frame guidance, our model effectively captures spatiotemporal correlations, as evidenced by the notably lower FVD$_{\text{I3D}}$ in \cref{tab2}: 1176.7 for our method versus 3557.7 for Imagen, 3513.5 for Stable Diffusion, and 1886.8 for Phenaki, demonstrating robust coherence across frames. While Phenaki offers some spatial consistency, it lacks the anatomical detail required for clinical use, whereas our approach achieves an optimal balance between resolution, computational efficiency, and anatomical fidelity.

Compared to GenerateCT, our method demonstrates notable improvements in anatomical fidelity, primarily due to anatomical priors provided by reports and masks, which guide the generation to produce images with structural integrity in the thoracic region and clearer abdominal organs, closely approximating real images. As shown in \cref{fig4}, GenerateCT exhibits substantial differences in scale across sagittal slices compared to our method and the ground truth, focusing only on the lung area, which leads to a loss of critical information and less clarity in abdominal organs. Additionally, in \cref{tab2}, GenerateCT’s performance in FID at 127.4 and CLIP score at 27.4 slightly lag behind ours, further validating our comprehensive advantages in anatomical accuracy, image quality, and semantic consistency. Noted that the negative FID value stems from highly similar slices.
\begin{table}
\centering
\tablestyle{8.0pt}{1.05}
\caption{Quantitative comparison of our method with baseline methods evaluated using the framework from GenerateCT.}
\vspace{-3mm}
\resizebox{1\columnwidth}{!}{
\begin{tabular}{c|cccc}
\textbf{Method} & \textbf{Out} & \textbf{FID}$\downarrow$ & \textbf{FVD$_{\text{I3D}}$}$\downarrow$ & \textbf{CLIP}$\uparrow$ \\
\shline
\rowcolor{darkgray}
GT/GT & - & -6.7&472.778&29.9\\
Base w/ Imagen \cite{saharia2022photorealistic} & 2D & 160.8 & 3557.7 & 24.8 \\
Base w/ SD \cite{rombach2022high}  & 2D & 151.7& 3513.5 & 23.5 \\
Base w/ Phenaki \cite{villegas2022phenaki}  & 3D & 104.3 & 1886.8 & 25.2 \\
Base w/ GenerateCT \cite{hamamci2023generatect} & 3D & 127.4 & 1382.4 & 27.4 \\
\rowcolor{baselinecolor}
\textbf{Ours} & 3D & \textbf{98.6} & \textbf{1176.7} & \textbf{29.4} \\
\Xhline{1pt}
\end{tabular}}
\label{tab2}

\end{table}

\subsection{Ablation Study}
To facilitate modular analysis, we abbreviated key components of TRACE as follows: (i) Paired Frame Modeling (PFM, Sec 3.2.1) vs. single-frame generation, (ii) Dual Anatomical Guidance (DAG, Sec 3.2.2) vs. unconditional generation, and (iii) Overlapping Frame Guidance (OFG, Sec 3.3) vs. Traditional Markovian inference. As depicted in \cref{tab1} (row 5-7), PFM improves background consistency and reduces flickering by leveraging inter-frame dependencies, while DAG enhances imaging quality with explicit anatomical priors. Their combination yields smoother motion but lacks long-range interactions, which OFG addresses by recursively fusing local and global cues for coherent anatomy.
\noindent\setlength{\columnsep}{3mm}\begin{wrapfigure}[8]{r}{0.4\columnwidth}
    \vspace{-4mm}
    \centering
    \includegraphics[height=0.235\columnwidth,width=0.38\columnwidth]{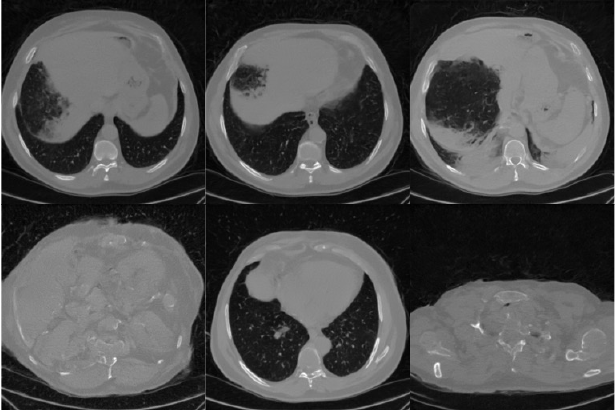}
    \vspace{-3mm}
\caption{\small Ablation results for anatomical mask granularity.}
    \label{fig5}
\end{wrapfigure}
\textbf{Anatomical Mask Granularity.} We further evaluated mask granularity by comparing (i) a two-class setup (all anatomical structures vs. background) as shown in \cref{fig5}, row 1, (ii) a 128-class detailed mask setup in \cref{fig5}, row 2, and (iii) a three-class setup (lungs, other organs/skeletal structures, background) shown in \cref{fig3}(a). The results indicate that the three-class setup provides the best consistency and quality. In contrast, the two-class setup reduces the separation between the lungs and other organs, leading to misclassifications (e.g., generating the spleen as lung tissue) and disrupting lung continuity. The 128-class setup performs well in the middle thorax region but fails to converge in the upper and lower thorax, resulting in incomplete or abnormal structures in these areas.

\noindent\textbf{Skipping Intervals and Sample Size.} To evaluate the impact of training sample size on model performance and to assess the role of frame skipping in capturing varying CT volume Z-axis spacing and enhancing spatial understanding, we designed this ablation study. We tested three skip intervals: (i) consecutive pairs (k = 1), (ii) skips of 1 and 2 frames (k = 1, 2), and (iii) skips of 1, 2, and 4 frames (k = 1, 2, 4), using two different training sizes (50 and 100 patients). As shown in \cref{tab1} (row 8-11), the combined skip interval (1, 2, 4) best captures temporal dependencies, particularly achieving the highest performance with the larger dataset (100 patients, row 11, ours).



\begin{table}
    \centering
    \tablestyle{8.0pt}{1.05}
    \caption{Expert Evaluation Scores for TRACE-Generated 3D Chest CT Volumes by Two Radiologists.}
    \vspace{-3mm}
    \resizebox{0.8\columnwidth}{!}{%
    \begin{tabular}{lcc}
        \textbf{Evaluation Criteria} & \textbf{Radiologist 1} & \textbf{Radiologist 2} \\
        \shline
        \rowcolor{gray!20}
        \multicolumn{3}{l}{\textbf{Anatomical Fidelity}} \\
        Structural Accuracy & 8.8 ± 0.9 & 8.5 ± 1.0 \\
        Tissue Contrast & 9.0 ± 0.8 & 8.9 ± 0.7 \\
        Pathological Representation & 7.8 ± 1.1 & 7.5 ± 1.2 \\
        \hline
        \rowcolor{gray!20}
        \multicolumn{3}{l}{\textbf{Spatiotemporal Consistency}} \\
        Intra-slice Consistency & 8.9 ± 0.7 & 9.1 ± 0.6 \\
        Inter-slice Continuity & 9.2 ± 0.6 & 9.0 ± 0.7 \\
        Temporal Artifacts & 8.7 ± 0.8 & 8.8 ± 0.8 \\
        \hline
        \rowcolor{gray!20}
        \multicolumn{3}{l}{\textbf{Overall Diagnostic Utility}} \\
        Overall Image Quality & 8.5 ± 1.0 & 8.7 ± 0.9 \\
        Confidence Level & 8.6 ± 0.9 & 8.8 ± 0.8 \\
        \Xhline{1pt}
    \end{tabular}%
    }
    \label{tab:expert_eval_reliability}

\end{table}

\subsection{Expert evaluation}
To rigorously assess the clinical applicability and reliability of TRACE, an expert evaluation was conducted involving two board-certified thoracic radiologists with 4 and 6 years of specialized experience, respectively. The radiologists independently evaluated fifty 3D chest CT volumes generated by TRACE against the ground truth, scoring them across three primary criteria: anatomical fidelity, spatiotemporal consistency, and overall diagnostic utility, using a 10-point scale. As summarized in \cref{tab:expert_eval_reliability}, TRACE demonstrated strong performance, achieving average scores of 8.7 for anatomical fidelity, 9.0 for spatiotemporal consistency, and 8.7 for diagnostic utility. These results highlight TRACE's ability to meet stringent quality standards, underscoring its potential for integration into clinical workflows.

\section{Conclusion}
In this work, we introduce TRACE, a framework for generating anatomically faithful 3D chest CT volumes with 2D multimodal-conditioned diffusion. TRACE effectively addresses key challenges in 3D CT generation, addressing anatomical fidelity, spatiotemporal coherence, adaptable axial resolution, and reduced computational costs. Our approach further enhances evaluation by incorporating a fidelity assessment of generated volumes. Experimental results, supported by expert evaluations, indicate that TRACE surpasses existing methods in fidelity and efficiency, highlighting its potential for clinical applications in data augmentation and personalized modeling.

{
    \small
    \bibliographystyle{ieeenat_fullname}
    \bibliography{main}
}


\end{document}